\documentclass[runningheads]{llncs}
\usepackage[T1]{fontenc}
\usepackage{graphicx}
\usepackage{algorithm}
\usepackage{algorithmic}
\usepackage{booktabs}
\usepackage{tikz}
\usepackage{xcolor}
\usepackage{amsmath}
\usepackage{amssymb}
\usetikzlibrary{shapes.geometric, arrows.meta, positioning, calc}

\begin{document}

\title{S.E.A.G.R: A Socially and Emotionally Aware Greeting Robot Framework with Dual-Layer Cultural and Affective Modulation}
\titlerunning{S.E.A.G.R}

\author{
    Sajjad Hussain\textsuperscript{1},
    Ranveer Bhura\textsuperscript{2},
    Amandip Dutta\textsuperscript{2},
    Shwetangshu Biswas\textsuperscript{3}
}
\authorrunning{S. Hussain, R. Bhura, A. Dutta, S. Biswas}

\institute{
\textsuperscript{1}School of Architecture, Technology and Engineering, University of Brighton, Brighton, UK \\
\textsuperscript{2}Department of Mechanical Engineering, National Institute of Technology, Silchar, India \\
\textsuperscript{3}Department of Electrical Engineering, National Institute of Technology, Silchar, India \\
}

\maketitle

\begin{abstract}
This paper presents SEAGR (Socially and Emotionally Aware Greeting Robot), a robotic greeting framework designed for human--robot interaction environments involving users from diverse cultural backgrounds and different emotional states. Since greeting behaviour strongly influences first impressions, user comfort, and trust, robots operating in public spaces must be able to interact in a socially appropriate and adaptive manner. However, many existing systems still rely on static greeting routines that do not account for cultural variation, emotional context, or interpersonal distance. SEAGR introduces a dual-layer modulation framework in which cultural identity determines the appropriate greeting type, while affective cues influence how that greeting is executed. The system combines context-aware cultural mapping, emotion-based gesture modulation, and proxemic regulation within a unified Sense--Think--Act architecture. A low-cost prototype is implemented using a USB camera, ultrasonic sensor, Arduino-controlled servos, and a laptop-based Python processing system. This work is presented as a system design and proof-of-concept; empirical validation through user studies is explicitly acknowledged as a current limitation and is identified as the primary direction for future work.
\end{abstract}

\section{Introduction}

Human--Robot Interaction (HRI) is an increasingly important research area as robots enter social environments such as museums, hospitals, airports, and exhibitions. In such settings, robots must interact with humans in socially appropriate and culturally acceptable ways \cite{dautenhahn2007social}. Among the various forms of interaction, greeting behaviour plays a critical role as it represents the first point of contact between humans and robots, significantly influencing user perception, trust, and willingness to engage further.

Designing socially appropriate greeting behaviour is genuinely difficult. Human greeting practices vary significantly across cultures: a bow is common in Japan, a Namaste gesture is widely used in India, and a handshake is typically expected in many Western contexts \cite{trovato2016cross}. Robots relying on static routines fail to adapt, producing interactions that feel awkward or offensive. Beyond cultural variation, emotional state shapes how a greeting should unfold. Humans adjust their behaviour depending on whether the other person appears relaxed, hurried, or distressed, yet most existing robotic greeting systems do not incorporate affective cues and produce rigid patterns that feel intrusive in real-world settings.

A third important dimension is proxemics --- the spatial distance people maintain during social interaction. Research in HRI has consistently shown that initiating interaction from an inappropriate distance undermines user comfort and acceptance even when all other aspects of the greeting are well designed \cite{walters2009proxemics}. Despite its importance, proxemic regulation is often treated as a separate engineering problem rather than an integrated component of a broader social greeting framework.

It is worth noting that human customer service workers typically use standardised greetings --- ``Hello, how can I help you?'' --- and adapt only after the customer provides context. The key motivation for cultural adaptation in robotics is not that standardised greetings fail entirely, but that robots lack the implicit social awareness, body language reading, and conversational flexibility that allows humans to adapt naturally after the initial contact. A robot that begins with a culturally informed greeting is more likely to establish initial rapport, particularly with users from cultures where greeting rituals carry significant social weight.

To address these interconnected challenges, this paper presents \textbf{S.E.A.G.R} (Socially and Emotionally Aware Greeting Robot), a framework that generates culturally appropriate and emotionally adaptive greeting behaviours in public social environments. The contributions of this work are as follows:

\begin{itemize}
\item A dual-layer greeting modulation framework separating cultural greeting selection from emotion-based gesture execution, with formal mathematical descriptions of both layers.
\item A socially aware interaction pipeline integrating cultural mapping, affective motion modulation, and proxemic regulation within a unified Sense--Think--Act architecture.
\item A proof-of-concept prototype using low-cost sensors and rule-based decision logic, with a structured pathway toward empirical evaluation.
\item A discussion of ethical considerations including facial detection consent, cultural parody risk, neurodivergent user needs, and the limitations of badge-based identification.
\end{itemize}

The remainder of this paper is organised as follows. Section~2 reviews related work. Section~3 describes the proposed SEAGR system architecture with formal models. Section~4 presents the hardware and software implementation. Section~5 discusses limitations. Section~6 presents conclusions and future work.

\section{Literature Review}

HRI has emerged as an interdisciplinary field combining robotics, AI, psychology, human factors, and social science \cite{goodrich2007human}. Dautenhahn \cite{dautenhahn2007social} established that robots must incorporate social intelligence to be accepted in everyday environments. Goodrich and Schultz \cite{goodrich2007human} surveyed key interaction dimensions including autonomy, communication modalities, and interaction architecture. Feil-Seifer and Mataric \cite{feilseifer2011socially} demonstrated that robots provide meaningful value through social interaction, and Kanda et al. \cite{kanda2004interactive} showed that combining embodiment, verbal interaction, and social presence enables long-term human--robot partnerships.

\subsection{Proxemics, Culture, and Emotion in HRI}

Proxemic behaviour is a core requirement for socially acceptable robot interaction. Walters et al. \cite{walters2009proxemics} showed that humans maintain different distances from robots than from other humans, requiring HRI-specific proxemic investigation. Takayama and Pantofaru \cite{takayama2009proxemics} demonstrated that a robot's motion behaviour strongly shapes user willingness to approach, while Mumm and Mutlu \cite{mumm2011proxemics} showed that distancing is psychologically mediated by trust. Samarakoon et al. \cite{samarakoon2022proxemics} emphasised that proxemic preferences depend on robot appearance, environment, and user characteristics, and Lehmann et al. \cite{lehmann2020proxemic} confirmed that personal space expectations reflect deeper psychological interpretations of the robot as a social entity. Lawrence et al. \cite{lawrence2025socialnorms} reinforced that socially acceptable robot behaviour is governed by broader norms around politeness, timing, and situational context.

Cultural differences similarly shape robot interaction. The CARESSES project \cite{bruno2017caresses} demonstrated how knowledge-based systems adapt speech and gestures according to user cultural background for elderly care. Papadopoulos et al. \cite{papadopoulos2020cultural} reported that culturally competent robots improve engagement in care settings, and Lim et al. \cite{lim2021culture} argued that culture must be a central HRI design parameter rather than an optional layer. Trovato et al. \cite{trovato2016cross} confirmed that culturally aligned robot greetings improve acceptance and reduce discomfort.

Affective adaptation is equally central. Breazeal \cite{breazeal2003emotion} established that emotional models make robots more understandable and socially responsive. Stock-Homburg \cite{stockhomburg2022emotions} confirmed across two decades of research that emotional expressiveness increases perceived social presence, trust, and acceptance. Kühnlenz et al. \cite{kuhnlenz2013helpfulness} showed that emotional adaptation positively influences user behaviour, and Rawal and Stock-Homburg \cite{rawal2022facial} highlighted the practical limitations of real-world emotion recognition, confirming that robust handling of ambiguous facial information remains an open challenge.

\subsection{Robotic Greeting Behaviour, Gesture, and Ethics}

Satake et al. \cite{satake2009approach} showed that appropriate greeting strategies including approach timing and spatial behaviour  significantly increase successful engagement in public environments. Urakami and Seaborn \cite{urakami2023nonverbal} argued that nonverbal cues are essential for social competence, and Mutlu et al. \cite{mutlu2009footing} demonstrated that gaze cues shape participant roles during robot conversations. A culturally correct greeting may still feel inappropriate if performed too abruptly or at an incorrect distance; robotic greeting must therefore be understood as a multimodal social act involving not only what greeting is selected, but also when, where, and how it is executed.

Recent work by Hussain et al. \cite{hussain2026gesture} demonstrated that lightweight computer vision pipelines achieve reliable real-time gesture execution without specialised hardware, a principle directly relevant to SEAGR's gesture primitives. Elendu et al. \cite{elendu2023ethical} highlighted the ethical dimensions of deploying AI in public environments  including consent, data privacy, and algorithmic bias  considerations that apply directly to systems involving facial scanning and cultural profiling.

Despite these advances, many existing studies treat cultural adaptation, emotion recognition, proxemic behaviour, and nonverbal communication as separate problems rather than integrating them into a unified framework. The present work is motivated by this gap and explicitly acknowledges the limitations of the current proof-of-concept while establishing a clear pathway toward empirical validation.

\section{Proposed System Architecture}

\begin{figure}[h]
\centering
\begin{tikzpicture}[
    node distance=1.3cm,
    every node/.style={font=\footnotesize},
    process/.style={rectangle, rounded corners=2pt, draw=blue!70, fill=blue!5, thick, minimum width=2.5cm, minimum height=0.7cm, text centered},
    decision/.style={diamond, draw=orange!70, fill=orange!10, thick, aspect=1.8, text centered, inner sep=0pt},
    start/.style={rectangle, rounded corners=10pt, draw=green!70!black, fill=green!10, thick, minimum width=2.5cm, minimum height=0.7cm, text centered},
    arrow/.style={->, >=stealth, thick, color=gray!80}
]
\node (start) [start] {User Approaches};
\node (social) [decision, below of=start] {Social Zone?};
\node (scan) [process, below of=social, yshift=-0.2cm] {Scan ID Badge};
\node (id) [decision, below of=scan, yshift=-0.2cm] {ID Found?};
\node (profile) [process, below left of=id, xshift=-1.2cm, yshift=-0.8cm] {Fetch Profile};
\node (neutral) [process, below right of=id, xshift=1.2cm, yshift=-0.8cm] {Generic Profile};
\node (select) [process, below of=id, yshift=-2.2cm] {Select Greeting Style};
\node (emotion) [process, below of=select] {Detect Emotion};
\node (execute) [start, below of=emotion] {Execute Response};
\draw [arrow] (start) -- (social);
\draw [arrow] (social) -- node[right, color=black]{Yes} (scan);
\draw [arrow] (social.west) -- ++(-0.5,0) |- node[pos=0.25, left, color=black]{No} (start.west);
\draw [arrow] (scan) -- (id);
\draw [arrow] (id.west) -| node[above, xshift=0.5cm, color=black]{Yes} (profile.north);
\draw [arrow] (id.east) -| node[above, xshift=-0.5cm, color=black]{No} (neutral.north);
\draw [arrow] (profile.south) |- (select.west);
\draw [arrow] (neutral.south) |- (select.east);
\draw [arrow] (select) -- (emotion);
\draw [arrow] (emotion) -- (execute);
\end{tikzpicture}
\caption{High-level system decision flow of the SEAGR framework.}
\label{fig:flowchart}
\end{figure}
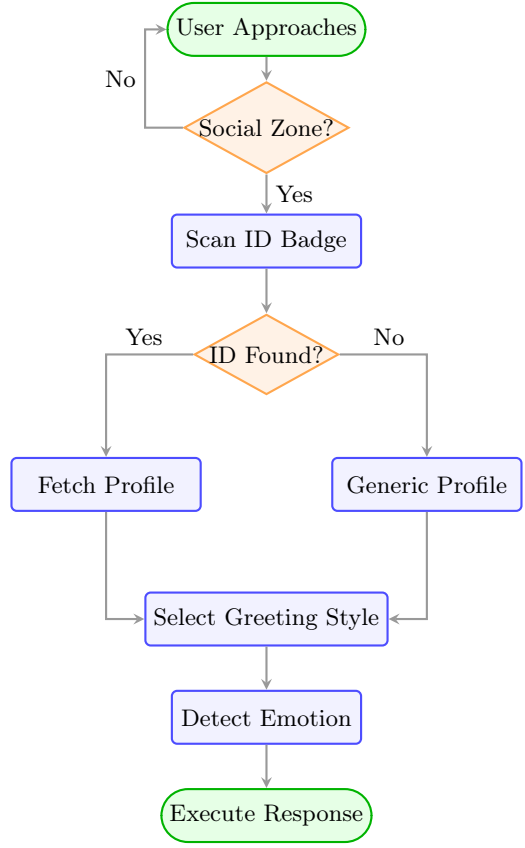

The proposed SEAGR system follows a Sense--Think--Act architecture that integrates perception, decision-making, and actuation to generate socially compliant greetings in real time. Fig.~\ref{fig:flowchart} illustrates the overall decision flow.

\subsection{Perception Module and Proxemic Gating}

The perception module detects user presence, measures interpersonal distance, and extracts basic affective cues. A USB camera identifies the user's badge and captures coarse facial expression information. An HC-SR04 ultrasonic sensor measures distance to enforce proxemic compliance. The module does not infer cultural identity from visual appearance, deliberately avoiding appearance-based profiling.

The current prototype consists of a laptop-based processing unit with servo-actuated upper-body components capable of head nodding, bowing, and basic arm gestures. It does not resemble a humanoid robot. Robot morphology is known to influence proxemic preferences and user comfort \cite{lehmann2020proxemic,samarakoon2022proxemics}, and the distance thresholds are calibrated for this specific form factor. Deployment on platforms of substantially different appearance would require recalibration.

The proxemic gate is defined formally as follows. Let $d$ denote the measured distance between the robot and the approaching user. The system defines two thresholds: $d_{\min}$ (minimum comfortable distance) and $d_{\max}$ (outer boundary of the social interaction zone). The interaction is activated only when:

\begin{equation}
d_{\min} \leq d \leq d_{\max}
\label{eq:proxemic_gate}
\end{equation}

In the current implementation, $d_{\min} = 0.5$\,m and $d_{\max} = 2.5$\,m, consistent with Hall's \cite{walters2009proxemics} social zone definition. If $d < d_{\min}$, the robot steps back; if $d > d_{\max}$, no interaction is initiated.

\subsection{Decision Logic Layer and Cultural Greeting Selection}

The decision logic layer is a rule-based framework ensuring predictable, explainable, and ethically compliant behaviour. When a user is detected within the social zone, the system attempts badge identification. The badge identifier is matched against a structured database of user-provided metadata including nationality and preferred language.

Formally, let $\mathcal{C} = \{c_1, c_2, \ldots, c_n\}$ be the set of cultural profiles stored in the attendee database, and let $b$ denote the scanned badge identifier. The cultural greeting selection function $\Phi$ is defined as:

\begin{equation}
\Phi(b) =
\begin{cases}
g_{c_i} & \text{if } b \mapsto c_i \in \mathcal{C} \\
g_{\text{neutral}} & \text{otherwise}
\end{cases}
\label{eq:greeting_selection}
\end{equation}

where $g_{c_i}$ is the culturally aligned greeting motor primitive associated with profile $c_i$, and $g_{\text{neutral}}$ is a default neutral greeting (head nod, verbal salutation, conservative distance) applied when no badge match is found. This fallback is consistent with HRI recommendations that robots should default to minimally intrusive behaviour when user context is unavailable \cite{lawrence2025socialnorms,satake2009approach}. Table~\ref{tab:greeting} shows the nationality-to-greeting mapping, informed by cross-cultural HRI literature \cite{trovato2016cross,lim2021culture}.

\begin{table}[h]
\caption{Nationality-Based Greeting Mapping.}
\label{tab:greeting}
\centering
\begin{tabular}{|l|l|l|}
\hline
\textbf{Nationality} & \textbf{Gesture} & \textbf{Greeting Phrase} \\
\hline
Japan    & Bow                  & ``Konnichiwa'' \\
India    & Namaste              & ``Namaste'' \\
USA      & Handshake Invitation & ``Hello'' \\
Germany  & Handshake Invitation & ``Guten Tag'' \\
France   & Nod                  & ``Bonjour'' \\
UAE      & Verbal Greeting      & ``As-salamu Alaikum'' \\
Brazil   & Handshake Invitation & ``Ol\'{a}'' \\
Thailand & Wai Gesture          & ``Sawasdee'' \\
Nigeria  & Nod                  & ``Hello'' \\
\hline
\end{tabular}
\end{table}

\subsection{Emotional Modulation Framework}

The emotional modulation framework adjusts how a greeting is performed based on the detected emotional state of the user. Crucially, this module does not change which greeting is selected  that is determined solely by Eq.~\ref{eq:greeting_selection}  but modifies the execution parameters of the already-selected greeting to improve perceived responsiveness and social sensitivity.

The system estimates coarse emotional states $e \in \mathcal{E} = \{\text{relaxed}, \text{neutral}, \text{stressed}, \text{unknown}\}$ from facial expression cues using broad categories to maintain robustness under real-world conditions. The final gesture amplitude is computed as:

\begin{equation}
G_{\text{final}} = \alpha(e) \times G_{\text{base}}
\label{eq:gesture_modulation}
\end{equation}

where $G_{\text{base}}$ is the predefined cultural gesture amplitude and $\alpha(e) \in (0, 1]$ is an emotion-dependent scaling factor. The scaling factors are defined as:

\begin{equation}
\alpha(e) =
\begin{cases}
1.00 & \text{if } e = \text{relaxed} \\
0.85 & \text{if } e = \text{neutral} \\
0.65 & \text{if } e = \text{stressed} \\
0.85 & \text{if } e = \text{unknown (safe default)}
\end{cases}
\label{eq:alpha}
\end{equation}

These values were informed by the proxemics and affective HRI literature \cite{takayama2009proxemics,kuhnlenz2013helpfulness} and represent conservative design choices that prioritise non-intrusiveness. A stressed user triggers reduced amplitude and shorter duration to avoid intrusiveness, while a relaxed user receives the greeting at full amplitude and normal timing. This approach is consistent with findings by Kühnlenz et al. \cite{kuhnlenz2013helpfulness} and Stock-Homburg \cite{stockhomburg2022emotions} that even coarse affective adaptation meaningfully improves user comfort and perceived social presence.

The gesture movement duration is also modulated according to:

\begin{equation}
T_{\text{final}} = \beta(e) \times T_{\text{base}}
\label{eq:duration}
\end{equation}

where $T_{\text{base}}$ is the nominal gesture duration and $\beta(e)$ follows the same mapping as $\alpha(e)$. The overall greeting response vector is therefore defined as:

\begin{equation}
\mathbf{R} = \left( \Phi(b),\; G_{\text{final}},\; T_{\text{final}},\; v_{\text{tone}} \right)
\label{eq:response_vector}
\end{equation}

where $v_{\text{tone}}$ is the vocal tone parameter (normal or softened) selected based on the detected emotional state. Algorithm~1 and Table~\ref{tab:emotion} summarise the gesture scaling procedure.

\begin{algorithm}
\caption{Emotion-Based Gesture Scaling}
\begin{algorithmic}
\STATE $G_{\text{base}} \leftarrow$ predefined cultural gesture amplitude
\STATE $T_{\text{base}} \leftarrow$ predefined gesture duration
\STATE Detect emotional state $e$ from facial expression cues
\STATE Retrieve $\alpha(e)$ and $\beta(e)$ from Eq.~\ref{eq:alpha}
\STATE $G_{\text{final}} \leftarrow \alpha(e) \times G_{\text{base}}$
\STATE $T_{\text{final}} \leftarrow \beta(e) \times T_{\text{base}}$
\STATE Execute greeting response vector $\mathbf{R} = (\Phi(b), G_{\text{final}}, T_{\text{final}}, v_{\text{tone}})$
\end{algorithmic}
\end{algorithm}

\begin{table}[h]
\caption{Emotion-Based Gesture Modulation Parameters.}
\label{tab:emotion}
\centering
\begin{tabular}{|l|c|c|l|}
\hline
\textbf{Detected Emotion} & \textbf{$\alpha(e)$} & \textbf{$\beta(e)$} & \textbf{Gesture Amplitude} \\
\hline
Happy / Relaxed  & 1.00 & 1.00 & Full \\
Neutral          & 0.85 & 0.85 & Moderate \\
Stressed / Sad   & 0.65 & 0.65 & Reduced \\
Unknown          & 0.85 & 0.85 & Moderate (safe default) \\
\hline
\end{tabular}
\end{table}

\subsection{Dual-Layer Modulation Overview}

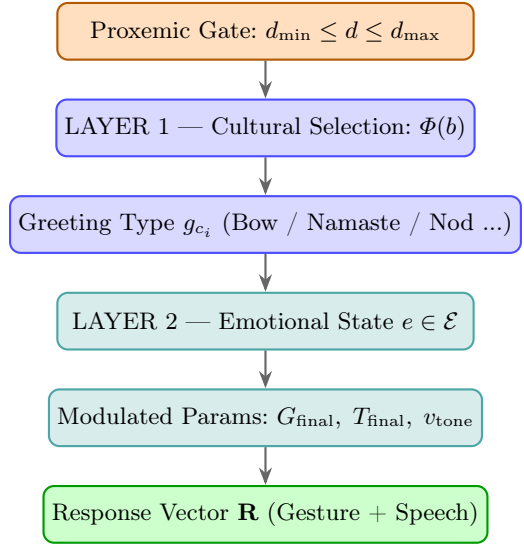
\begin{figure}[h]
\centering
\begin{tikzpicture}[
    node distance=0.55cm,
    box/.style={rectangle, rounded corners=4pt, draw, thick,
                minimum width=5.5cm, minimum height=0.75cm,
                text centered, font=\small},
    gate/.style={box,   fill=orange!25, draw=orange!70!black},
    layer1/.style={box, fill=blue!15,   draw=blue!70},
    layer2/.style={box, fill=teal!15,   draw=teal!70},
    outputbox/.style={box, fill=green!20, draw=green!60!black},
    arrow/.style={->, >=Stealth, thick, color=black!60}
]
\node (prox) [gate]                          {Proxemic Gate: $d_{\min} \leq d \leq d_{\max}$};
\node (l1)   [layer1, below=0.5cm of prox]  {LAYER 1 --- Cultural Selection: $\Phi(b)$};
\node (l1b)  [layer1, below=0.5cm of l1]    {Greeting Type $g_{c_i}$ (Bow / Namaste / Nod ...)};
\node (l2)   [layer2, below=0.5cm of l1b]   {LAYER 2 --- Emotional State $e \in \mathcal{E}$};
\node (l2b)  [layer2, below=0.5cm of l2]    {Modulated Params: $G_{\text{final}},\; T_{\text{final}},\; v_{\text{tone}}$};
\node (out)  [outputbox, below=0.5cm of l2b]   {Response Vector $\mathbf{R}$ (Gesture + Speech)};
\draw[arrow] (prox) -- (l1);
\draw[arrow] (l1)   -- (l1b);
\draw[arrow] (l1b)  -- (l2);
\draw[arrow] (l2)   -- (l2b);
\draw[arrow] (l2b)  -- (out);
\end{tikzpicture}
\caption{Dual-layer modulation pipeline of SEAGR. The proxemic gate enforces Eq.~\ref{eq:proxemic_gate}; Layer~1 applies the cultural selection function $\Phi(b)$; Layer~2 modulates execution parameters using Eqs.~\ref{eq:gesture_modulation}--\ref{eq:alpha}.}
\label{fig:duallayer}
\end{figure}

Fig.~\ref{fig:duallayer} summarises the dual-layer modulation principle. The proxemic gate (Eq.~\ref{eq:proxemic_gate}) activates the pipeline only when the user is within a socially appropriate distance. Layer~1 selects the greeting type via $\Phi(b)$ (Eq.~\ref{eq:greeting_selection}), and Layer~2 modulates execution via Eqs.~\ref{eq:gesture_modulation}--\ref{eq:response_vector}.

\section{Hardware and Software Implementation}

\begin{figure}[h]
\centering
\begin{tikzpicture}[
    node distance=1cm and 1.5cm,
    base/.style={rectangle, draw, thick, align=center, minimum height=1cm, minimum width=3cm, font=\small},
    sens/.style={base, fill=cyan!10, draw=cyan!60!black, rounded corners=2pt},
    proc/.style={base, fill=blue!10, draw=blue!70, rounded corners=5pt, minimum width=4cm},
    ctrl/.style={base, fill=orange!10, draw=orange!70, rounded corners=2pt},
    actu/.style={base, fill=purple!10, draw=purple!70!black, rounded corners=2pt},
    line/.style={draw, -{Stealth[scale=1.2]}, thick, color=gray!80}
]
\node (camera) [sens, xshift=-2cm] {\textbf{USB Camera} (Visual Input)};
\node (ultra)  [sens, xshift=3cm]  {\textbf{Ultrasonic} (HC-SR04)};
\node (laptop) [proc, below=1.2cm of $(camera.south)!0.5!(ultra.south)$] {
    \textbf{Main Laptop Controller} \\
    \scriptsize (Perception, Decision Logic \& CV)};
\node (arduino) [ctrl, below=1.2cm of laptop] {
    \textbf{Arduino Microcontroller} \\
    \scriptsize (PWM \& Signal Routing)};
\node (servo) [actu, below=1.2cm of arduino, xshift=-2.5cm] {\textbf{Servo Motors} \\ \scriptsize (Gestures)};
\node (audio) [actu, below=1.2cm of arduino, xshift=2.5cm]  {\textbf{Audio Speaker} \\ \scriptsize (Speech Synthesis)};
\draw [line] (camera.south) -- ++(0,-0.4) -| (laptop.north);
\draw [line] (ultra.south)  -- ++(0,-0.4) -| (laptop.north);
\draw [line] (laptop) -- (arduino) node[midway, right, font=\tiny, color=black]{Serial (USB)};
\draw [line] (arduino.south) -- ++(0,-0.4) -| (servo.north);
\draw [line] (arduino.south) -- ++(0,-0.4) -| (audio.north);
\node[left=1.5cm of camera,  font=\bfseries\color{gray}]{SENSE};
\node[left=1cm of laptop,    font=\bfseries\color{gray}]{THINK};
\node[left=1cm of arduino,   font=\bfseries\color{gray}]{CONTROL};
\node[left=0.5cm of servo,   font=\bfseries\color{gray}]{ACT};
\end{tikzpicture}
\caption{Hardware architecture of the SEAGR robotic greeting system.}
\label{fig:hardwareflow}
\end{figure}
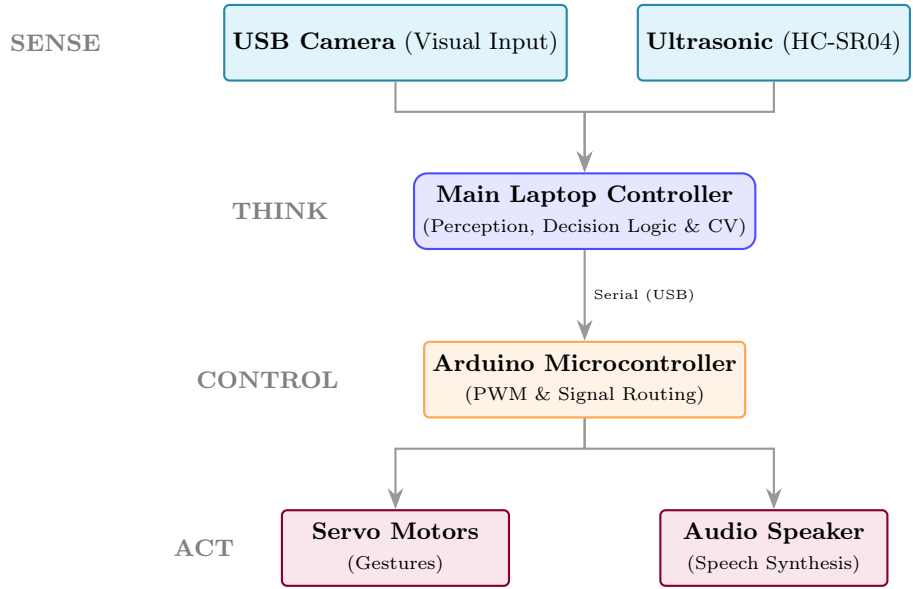

The SEAGR prototype uses a modular architecture integrating low-cost sensing devices, an embedded microcontroller, and a laptop-based perception and decision system. The system consists of three layers  Perception, Decision Logic, and Actuation  operating together as a complete interaction pipeline. The hardware architecture is illustrated in Fig.~\ref{fig:hardwareflow}.

\subsection{Sensing and Perception Subsystem}

The perception subsystem uses a USB camera and an HC-SR04 ultrasonic sensor. The camera performs face detection, badge recognition, and coarse facial expression analysis using the OpenCV Haar Cascade classifier, which identifies faces efficiently on standard hardware without a GPU. A Dlib 68-point landmark detector then extracts facial cues including eye openness, eyebrow position, and mouth shape. A lightweight FER2013-trained classifier estimates the user's emotional state, mapped to the three broad SEAGR categories defined in Eq.~\ref{eq:alpha}. The ultrasonic sensor provides reliable real-time proxemic distance measurement, compensating for the instability of purely vision-based distance estimation under variable lighting. Table~\ref{tab:sensors} summarises the sensing components.

\begin{table}[h]
\caption{Sensor Components Used in the System.}
\label{tab:sensors}
\centering
\begin{tabular}{|l|l|l|}
\hline
\textbf{Component} & \textbf{Purpose} & \textbf{Key Specification} \\
\hline
USB Camera & Face detection and badge scanning & 720p / 1080p resolution \\
HC-SR04 Ultrasonic Sensor & Distance measurement ($d$ in Eq.~\ref{eq:proxemic_gate}) & Range: 2\,cm -- 400\,cm \\
\hline
\end{tabular}
\end{table}

\subsection{Actuation Subsystem and Software Architecture}

Robot gestures are executed using servo motors connected to the Arduino microcontroller. The mechanical structure supports upper-body movements including head nodding, bowing, and the Namaste gesture, implemented as predefined motion primitives to ensure smooth and repeatable execution. The Arduino converts high-level gesture commands from the laptop into servo angle positions and timing sequences locally, reducing processing load. An audio speaker delivers verbal greeting output in the appropriate language, producing multimodal interaction. This design is supported by Urakami and Seaborn \cite{urakami2023nonverbal} and Kanda et al. \cite{kanda2004interactive}, who showed that combining gesture, voice, and embodiment significantly improves perceived social competence. Tables~\ref{tab:actuators} and~\ref{tab:software} list the actuation components and software stack respectively.

\begin{table}[h]
\caption{Actuation Components.}
\label{tab:actuators}
\centering
\begin{tabular}{|l|l|l|}
\hline
\textbf{Component} & \textbf{Function} & \textbf{Interface} \\
\hline
Arduino Uno   & Servo motor control   & Serial USB  \\
Servo Motors  & Gesture execution ($G_{\text{final}}$) & PWM control \\
Audio Speaker & Voice greeting output & Audio jack / USB \\
\hline
\end{tabular}
\end{table}

\begin{table}[h]
\caption{Software Stack Used in the System.}
\label{tab:software}
\centering
\begin{tabular}{|l|l|}
\hline
\textbf{Software Component} & \textbf{Function} \\
\hline
Python                  & Main system implementation \\
OpenCV                  & Face detection and image processing \\
PySerial                & Laptop-Arduino communication \\
Arduino IDE             & Servo motor control firmware \\
JSON / CSV Database     & Attendee metadata ($\mathcal{C}$) storage \\
Text-to-Speech Engine   & Verbal greeting generation \\
\hline
\end{tabular}
\end{table}

Attendee metadata is stored in structured JSON or CSV files. When a badge is recognised, the system retrieves the corresponding cultural profile and evaluates $\Phi(b)$ (Eq.~\ref{eq:greeting_selection}). The laptop communicates with the Arduino via serial interface, sending symbolic gesture commands (\texttt{G1} Namaste, \texttt{G2} bow, \texttt{G3} handshake, \texttt{G4} head nod). Gesture and speech are synchronised to produce a complete greeting act, with amplitude and timing determined by $\mathbf{R}$ (Eq.~\ref{eq:response_vector}). Prior work by Hussain et al. \cite{hussain2026gesture} demonstrated that similarly lightweight pipelines achieve reliable real-time gesture execution for robotic arm control, supporting the feasibility of this approach in time-constrained interaction scenarios.

\section{Limitations}

While the SEAGR framework presents a coherent and interpretable system design, several important limitations must be explicitly acknowledged.

\subsection{Empirical Validation and Badge-Based Identification}

The most significant limitation is the absence of user studies or subjective evaluations. Claims regarding social appropriateness, user comfort, and adaptive effectiveness are grounded in design reasoning and prior literature rather than experimental evidence; the framework cannot yet be considered validated. Future work must conduct user studies across diverse cultural groups, measuring perceived comfort, trust, naturalness, and social appropriateness against a standardised greeting baseline. The badge-scanning mechanism is well-suited to bounded conference environments but does not generalise to museums or public service spaces. A promising extension is a two-step interaction sequence in which the robot begins with a neutral greeting and then adapts based on the user's verbal or gestural response, preserving the ethical commitment to avoiding appearance-based cultural inference while broadening applicability.

\subsection{Facial Expression Limitations and Privacy}

Coarse emotion recognition is inherently uncertain in real-world conditions. The system cannot reliably distinguish a genuinely distressed user from someone whose neutral resting expression appears tense to the FER2013 classifier --- a well-documented perceptual limitation in automated emotion recognition systems. To mitigate this, the framework uses only three broad emotional categories and defaults to the neutral profile ($\alpha = 0.85$) when confidence is low, prioritising safety over expressiveness. Facial scanning also raises important ethical considerations. Entering a public space should not constitute implied consent to facial detection. Any real-world deployment must include clear prior notification, explicit opt-in consent mechanisms, and the option to interact without facial scanning, aligning with growing ethical frameworks for AI in public spaces \cite{elendu2023ethical}.

\subsection{Environmental Contexts, Cultural Parody, and Parameter Justification}

Standard proxemic rules may break down in noisy or crowded environments, and neurodivergent users may adjust interaction distance in ways that conflict with the robot's thresholds defined in Eq.~\ref{eq:proxemic_gate}. Future iterations should incorporate environmental density awareness and allow users to override distance regulation. A non-humanoid robot performing culturally specific gestures  such as a bow or Namaste also risks being perceived as patronising rather than respectful. The current system maps nationality to a default greeting type without asking users how they wish to be greeted by a robot specifically. Future versions should allow users to specify greeting preferences during registration, reducing reliance on cultural generalisation. Finally, the scaling factors $\alpha$ and $\beta$ (Eq.~\ref{eq:alpha}) and the proxemic thresholds $d_{\min}$, $d_{\max}$ were determined through iterative design reasoning informed by proxemics literature \cite{takayama2009proxemics,walters2009proxemics} rather than empirical optimisation, and must be validated and refined through controlled user studies before they can be treated as principled design choices.

\section{Conclusion and Future Work}

This paper presented SEAGR, a socially and emotionally aware robotic greeting framework designed as a system design and proof-of-concept for generating culturally appropriate and context-sensitive greeting behaviours in HRI environments. The system introduces a dual-layer modulation approach with formal mathematical descriptions: cultural greeting selection via the function $\Phi(b)$ (Eq.~\ref{eq:greeting_selection}), and emotion-based execution modulation via the response vector $\mathbf{R}$ (Eq.~\ref{eq:response_vector}), integrated with a proxemic gating condition (Eq.~\ref{eq:proxemic_gate}) within a unified Sense--Think--Act architecture. The framework enables robots to produce socially compliant and adaptive greeting behaviours using accessible, low-cost hardware and lightweight software tools.

It is explicitly acknowledged that the current work has not been validated through user studies. The claims made throughout this paper are grounded in design reasoning and the supporting literature rather than empirical measurement. This is the primary limitation of the work and is recognised as the necessary next step before the framework's effectiveness can be substantiated  not merely deferred as future work.

Future work will conduct user studies across diverse cultural groups measuring perceived comfort, social appropriateness, trust, and engagement against standardised greeting baselines. Ethical protocols for facial scanning consent will be developed from the outset. The research will also explore a two-step adaptive interaction model for non-badged environments, empirical calibration of the heuristic parameters $\alpha$, $\beta$, $d_{\min}$, and $d_{\max}$, and more robust emotion recognition incorporating speech and gaze cues. These developments will progressively move SEAGR from a conceptually grounded proof-of-concept toward a rigorously evaluated and deployable system for multicultural public interaction environments.

\bibliographystyle{splncs04}
\bibliography{references}

\end{document}